\documentclass{amsart}

\usepackage{amsmath,amssymb,amsthm,a4wide,caption,siunitx,hyperref}
\usepackage[ruled,linesnumbered]{algorithm2e}
\usepackage{mathtools}

\usepackage{natbib}
\usepackage{pgfplots}

\usepackage{graphicx,tikz}
\usepackage{cleveref}
\newtheorem{theorem}{Theorem}

\newcommand{\bR}{\mathbb{R}}
\newcommand{\eps}{\varepsilon}

\newcommand{\bbmat}{\begin{bmatrix}}
\newcommand{\ebmat}{\end{bmatrix}}
\newcommand{\beq}{\begin{equation}}
\newcommand{\eeq}{\end{equation}}
\newcommand{\tp}{\tilde{\phi}}

\newcommand{\ty}{\tilde{y}}
\newcommand{\tz}{\tilde{z}}
\newcommand{\tK}{\tilde{K}}
\newcommand{\ymin}{y_{\mathrm{min}}}
\newcommand{\ymax}{y_{\mathrm{max}}}

\theoremstyle{definition}

\theoremstyle{remark}
\newtheorem{remark}[theorem]{Remark}

\newcommand{\Nint}{N_{\text{int}}}

\makeatletter
\def\@setthanks{\vspace{-\baselineskip}\def\thanks##1{\@par##1\@addpunct.}\thankses}
\makeatother

\begin{document}

\title[]{Efficient Algorithms for t-distributed Stochastic Neighborhood Embedding}

\author[]{George C. Linderman}
\thanks{GCL was supported in part by NIH grant  \#1R01HG008383-01A1 and U.S. NIH MSTP Training Grant T32GM007205,
MR was supported in part by AFOSR grant \# FA9550-16-10175 and NIH grant \#1R01HG008383-01A1,
and YK was supported in part by NIH grant \#1R01HG008383-01A1.}
\address{Applied Mathematics Program, Yale University, New Haven, CT 06511, USA}
\email{george.linderman@yale.edu}

\author[]{Manas Rachh}
\address{Applied Mathematics Program, Yale University, New Haven, CT 06511, USA}
\email{manas.rachh@yale.edu}

\author[]{Jeremy G. Hoskins}
\address{Applied Mathematics Program, Yale University, New Haven, CT 06511, USA}
\email{jeremy.hoskins@yale.edu}

\author[]{\\Stefan Steinerberger}
\address{Department of Mathematics, Yale University, New Haven, CT 06511, USA}
\email{stefan.steinerberger@yale.edu}

\author[]{Yuval Kluger}
\address{\parindent0pt Department of Pathology and Applied Mathematics Program, Yale University School of Medicine, \-\ \-\ \-\ \-\ New Haven, CT 06511 USA}
\email{yuval.kluger@yale.edu}

\begin{abstract} 
	t-distributed Stochastic Neighborhood Embedding (t-SNE) is a method for
	dimensionality reduction and visualization that has become widely
	popular in recent years. Efficient implementations of t-SNE are
	available, but they scale poorly to datasets with hundreds of thousands
	to millions of high dimensional data-points. We present Fast Fourier
	Transform-accelerated Interpolation-based t-SNE (FIt-SNE), which
	dramatically accelerates the computation of t-SNE. The most
	time-consuming step of t-SNE is a convolution that we accelerate by
	interpolating onto an equispaced grid and subsequently using the fast
	Fourier transform to perform the convolution. We also optimize the
	computation of input similarities in high dimensions using
	multi-threaded approximate nearest neighbors. We further present a
	modification to t-SNE called ``late exaggeration,'' which allows for
	easier identification of clusters in t-SNE embeddings.  Finally, for
	datasets that cannot be loaded into the memory, we present out-of-core
	randomized principal component analysis (oocPCA), so that the top principal
	components of a dataset can be computed without ever fully loading the
	matrix, hence allowing for t-SNE of large datasets to be computed on
	resource-limited machines.
\end{abstract}

\maketitle

\section{Introduction}
In many fields, the visualization of large, high-dimensional datasets is
essential. t-distributed Stochastic Neighborhood Embedding (t-SNE), introduced
by \cite{maaten2008visualizing}, has become enormously popular in many fields,
such as in the analysis of single-cell RNA-sequencing (scRNA-seq) data, where
it is used to discover the subpopulations among large numbers of cells in an
unsupervised fashion. Unfortunately, even efficient methods for approximate
t-SNE require many hours to embed datasets on the order of hundreds of
thousands to millions of points, as often encountered in scRNA-seq and
elsewhere.  In this paper, we present Fast Fourier Transform-accelerated
Interpolation-based t-SNE (FIt-SNE) for fast and accurate computation of t-SNE,
essentially making it feasible to use t-SNE on datasets of this scale.
Furthermore, we build on recent theoretical advances to more clearly separate
clusters in t-SNE embeddings. Finally, we present an out-of-core implementation
of randomized principal component analysis (oocPCA) so that users can embed datasets
that are too large to load in the memory.
\subsection{t-distributed Stochastic Neighborhood Embedding}
Given a $d$-dimensional dataset $X = \{x_1, x_2, ..., x_N\} \subset
\mathbb{R}^d$, t-SNE aims to compute the low dimensional embedding $Y = \{y_1,
y_2, ..., y_N \} \subset \mathbb{R}^s$ where $s\ll d$, such that if two points
$x_i$ and $x_j$ are close in the input space, then their corresponding points
$y_i$ and $y_j$ are also close.  Affinities between points $x_i$ and $x_j$ in the input space, $p_{ij}$, are defined as

$$p_{i|j} = \frac{\exp{(-\|  x_i -  x_j \|^2 /2 \sigma_i^2 )}}{\sum_{k\neq i} \exp{( - \|  x_i -  x_k \|^2/ 2 \sigma_i^2 }) } \qquad \mbox{and} \qquad \label{p_ij}
	p_{ij}  = \frac{ p_{i|j} + p_{j|i} }{2N}.$$

$\sigma_i$ is the bandwidth of the Gaussian distribution, and it is chosen
using such that the perplexity of $P_i$ matches a given value, where $P_i$ is
the conditional distribution of all the other points given $x_i$. Similarly,
the affinity between points $y_i$ and $y_j$ in the embedding space is defined
using the Cauchy kernel

$$q_{ij} = \frac{(1 + \| y_i -  y_j\|^2)^{-1}}{\sum_{k\neq l} (1 +\|  y_k -  y_l\|^2 )^{-1}}.$$

t-SNE finds the points $\{y_1,..., y_n\}$ that minimize the Kullback-Leibler divergence between the joint distribution of points in the input space $P$ and the joint distribution of the points in the embedding space $Q$,

$$C(\mathcal Y) = KL(P || Q) = \sum_{i\neq j} p_{ij} \log \frac{p_{ij}}{q_{ij}}.$$

Starting with a random initialization, the cost function $C(\mathcal Y)$ is minimized by gradient descent, with the gradient (as derived by \cite{maaten2008visualizing})

$$\frac{\partial C}{\partial  y_i } = 4 \sum\limits_{j \neq i} (p_{ij} - q_{ij})q_{ij} Z ( y_i -  y_j),$$
where $Z$ is a global normalization constant
$$Z = \sum_{k\neq l}{ \left( 1 + \| y_k -
 y_l \|^2\right)^{-1}}.$$ 

We split the gradient into two parts

$$	\frac{1}{4} \frac{\partial C}{\partial  y_i } =  \sum_{j\neq i} {p_{ij} q_{ij} Z (  y_i -  y_j)} - \sum_{j\neq i}{ q_{ij}^2 Z ( y_i - y_j)} $$

where the first sum $F_{\text{attr},i}$ corresponds to an attractive force between points and the
second sum $F_{\text{rep},i}$ corresponds to a repulsive force

$$	\frac{1}{4} \frac{\partial C}{\partial  y_i } =   F_{\text{attr},i} - F_{\text{rep},i}. $$

The computation of the gradient at each step is an $n$-body simulation, where
the position of each point is determined by the forces exerted on it by all
other points. Exact computation of $n$-body simulations scales as $O(n^2)$,
making exact t-SNE computationally prohibitive for datasets with tens of
thousands of points. 
Accordingly, \cite{van2014accelerating}'s popular
implementation of t-SNE produces an approximate solution, and can be used on
larger datasets. In that implementation, they approximate $F_{\text{attr},i}$
 by nearest neighbors as computed using vantage-point trees (\cite{yianilos1993data}). Since the input
similarities do not change, they can be precomputed, and hence do not dominate
the computational time.  On the other hand, the repulsive forces
$F_{\text{rep},i}$ are approximated at each iteration using the Barnes-Hut
Algorithm (\cite{barnes1986hierarchical}), a tree-based algorithm which scales
as $O(n\log n)$. 
Despite these accelerations, it can still take many hours to run t-SNE on large scRNA-seq
datasets. Furthermore, given that t-SNE is often run many times with different
initializations to find the best embedding, faster algorithms are needed.
In this work, we present an approximate nearest neighbor based implementation
for computing $F_{\text{attr},i}$ and an interpolation-based fast Fourier
transform accelerated algorithm for computing $F_{\text{repul},i}$, both of
which are significantly faster than current methods.

\subsection{Early exaggeration}

\cite{van2014accelerating} and \cite{maaten2008visualizing} note that as the
number of points $n$ increases, the convergence rate slows down. To circumvent
this problem, implementations of t-SNE multiply the $F_{\text{attr},i}$ term by a
constant $\alpha >1$ during the first $250$ iterations of gradient descent:

$$	\frac{1}{4} \frac{\partial C}{\partial  y_i } =   \alpha F_{\text{attr},i} - F_{\text{rep},i} $$

This ``early exaggeration'' forces the points into tight clusters which can
move more easily, and are hence less likely to get trapped in local minima.
\cite{linderman2017clustering} showed that this early exaggeration phase is
essential for convergence of the algorithm and that when the exaggeration
coefficient $\alpha$ is set optimally, t-SNE is guaranteed to recover
well-separated clusters. In Section \S \ref{sec:exaggeration} we show that late
exaggeration (i.e. setting $\alpha >1$ during the last several hundred
iterations) is also useful, and can result in improved separation of clusters.
 
\subsection{Organization}
The organization of this paper is as follows: we first present and benchmark a
fast Fourier transform accelerated interpolation-based method for optimizing
the computation of $F_{\text{rep},i}$ in Section \S \ref{sec:repul}.
Section \S \ref{sec:attr} describes methods for accelerating the computation of input
similarities $p_{ij}$ required for $F_{\text{attr},i}$. Section \S
\ref{sec:exaggeration} describes ``late exaggeration'' for improving separation
of clusters in t-SNE embeddings.  Section \S \ref{sec:heatmaps} describes t-SNE
heatmaps, an application of 1-dimensional t-SNE to the visualization of
single-cell RNA-sequencing data. Finally, Section \S \ref{sec:oocPCA} presents
an implementation of out-of-core PCA for the analysis of datasets too large to
fit in the memory.
\section{The repulsive forces $F_{\text{rep},i}$}
\label{sec:repul}
Suppose $\{y_{1},y_{2},\dots,y_{N}\}$ is an $s$-dimensional embedding of a collection of $d$-dimensional vectors $\{x_1,\dots,x_N\}.$ 
At each step of gradient descent, the repulsive forces are given by
\beq
\label{eq:frep}
F_{\text{rep},k}(m) = 
\left(\sum_{\substack{\ell=1 \\ \ell \neq k}}^{N} 
\frac{y_\ell(m) - y_k(m)}
{\left(1 + \| y_\ell - y_k\|^2\right)^{2}} \right) \Bigg/
\left(
\mathop{
\sum_{\substack{j = 1}}^{N}
\sum_{\substack{\ell = 1}}^{N}}_{\ell \neq j}
\frac{1}{(1 + \| y_\ell - y_j\|^2)}\right) 
  \, , 
\eeq  
where $k=1,2,\ldots N$, $m=1,2\ldots s,$ and $y_i(j)$ denotes the $j^{\rm th}$ component of $y_i.$ Evidently, the repulsive force between
the vectors $\{y_1,\dots,y_N\}$ consists of $N^2$ pairwise interactions, and were it computed directly, would require CPU-time scaling as $O(N^2).$ Even for datasets consisting of a few thousand points, this cost becomes prohibitively expensive. 
Our approach enables the accurate computation of these pairwise interactions in $O(N)$ time. 
Since the majority of applications of t-SNE are for two-dimensional embeddings (and in \S\ref{sec:heatmaps} we present an application of one-dimensional embeddings), in the following we focus our attention on the cases where $s=1$ or $2.$ However, we note that our algorithm extends naturally to arbitrary dimensions. In such cases, though the constants in the computational cost will vary, our approach will still yield an algorithm with a CPU-time which scales as $O(N).$

We begin by observing that the repulsive forces $F_{\text{rep},k}$ defined in~\cref{eq:frep} 
can be expressed as $s+2$ 
sums of the form
\beq
\label{eq:nbody2}
\phi(y_{i}) = \sum_{j=1}^{N} K(y_{i},y_{j}) q_{j} 
\eeq
where the kernel $K(y,z)$ is either
\begin{equation}
\label{eq:kers}
K_{1}(y,z) = \frac{1}{(1+\| y-z \|^2)} \, , \quad \text{or} \quad K_{2}(y,z) = \frac{1}{(1+\|y-z \|^2)^2} \, , 
\end {equation}
	for $y,z \in \bR^{s}$ (see Appendix). 
Note that both of the kernels $K_{1}$ and $K_{2}$ are smooth functions of
$y,z$ for all $y,z \in \bR^{s}$.
The key idea of our approach is to use polynomial interpolants of
the kernel $K$ in order to accelerate the evaluation of the
$N-$body interactions defined in~\cref{eq:nbody2}.

\subsection{Mathematical Preliminaries}
\label{subsec:repelprelim}

First, we demonstrate with a simple example how polynomial interpolation can be used to accelerate the computation
of the $N-$body interactions with a smooth kernel.
Suppose that $y_1,\dots,y_M \in (y_0,y_0+R)$ and  $z_1,\dots,z_N \in (z_0,z_0+R)$. 
Let $I_{y_{0}}$ and $I_{z_{0}}$ denote the intervals  $(y_0,y_0+R)$ and $ (z_0,z_0+R)$, respectively.
Note that no assumptions are made regarding the relative locations of $y_0$ and $z_0;$ in particular, the case $y_0 = z_0$ is also permitted.

Now consider the sums
\beq
\label{eq:phi1}
\phi(y_{i}) = \sum_{j=1}^{N} K(y_{i},z_{j}) q_{j}  \, , \quad i=1,2,\ldots M \, .
\eeq

Let $p$ be a positive integer. 
Suppose that $\tz_1,\dots,\tz_p,$ are a collection of $p$
points on the interval $I_{z_0}$ and that 
$\ty_1,\dots,\ty_p$,  are a collection
of $p$ points on the interval $I_{y_0}$.
Let $K_{p}(y,z)$
denote a bivariate polynomial interpolant of the kernel $K(y,z)$
satisfying
\beq
\nonumber
K_{p}(\ty_{j},\tz_{\ell}) = K(\ty_{j},\tz_{\ell})\,,  \quad
j,\ell=1,2,\ldots p.
\eeq
A simple calculation shows that $K_{p}(y,z)$ is given by
\beq
\label{eq:kerinterp}
K_{p}(y,z) = \sum_{\ell=1}^{p} \sum_{j=1}^{p} K(\ty_{j},\tz_{\ell}) 
L_{j,\ty}(y) L_{\ell,\tz}(z) \, ,
\eeq
where $L_{j,\ty}(y)$ and 
$L_{\ell,\tz}(z)$ are the Lagrange polynomials 
\begin{equation}
\nonumber
L_{\ell,\ty}(y) = 
\prod_{\substack{j=1\\j\neq \ell}}^{p} (y-\ty_{j}) \Big/
\prod_{\substack{j=1\\j\neq \ell}}^{p} (\ty_{\ell}-\ty_{j}) \, , \quad
\text{and} \quad 
L_{\ell,\tz}(z) = 
\prod_{\substack{j=1\\j\neq \ell}}^{p} (z-\tz_{j}) \Big/
\prod_{\substack{j=1\\j\neq \ell}}^{p} (\tz_{\ell}-\tz_{j}) \, ,
\end{equation}
$\ell=1,2\ldots p$. In the following we will refer to the points $\ty_1,\dots,\ty_p$, and $\tz_1,\dots,\tz_p$ as interpolation points. 

Let $\tp(y_{i})$ denote the approximation to $\phi(y_{i})$ obtained by replacing the kernel $K$ in ~\cref{eq:phi1}  by its polynomial
interpolant $K_{p}$, i.e.
\beq
\nonumber
\tp(y_{i}) = \sum_{j=1}^{N} K_{p}(y_{i},z_{j}) q_{j} \, ,
\eeq
for $i=1,2\ldots M$. 
Clearly the error in approximating $\phi(y_{i})$ 
via $\tp(y_{i})$ is bounded (up to a constant) by the error
in approximating $K(y,z)$ via $K_{p}(y,z)$. In particular,
if the polynomial interpolant satisfies the
inequality
\beq
\nonumber
\sup_{\substack{y\in(y_{0},y_{0}+R)\\ z \in (z_{0},z_{0}+R)}}
|K_{p}(y,z)-K(y,z)| \leq \eps \,, 
\eeq
then the error $|\tp(y_{i})-\phi(y_{i})|$ is given by
\begin{align*}
|\tp(y_{i}) - \phi(y_{i})|
&=\left| \sum_{j=1}^{N} \left( K_{p}(y_{i},z_{j}) - K(y_{i},z_{j}) \right) q_{j}\right| \\
&\leq \sum_{j=1}^{N} \left| K_{p}(y_{i},z_{j}) - K(y_{i},z_{j}) \right| |q_{j}| \\
&\leq \eps \sum_{j=1}^{N} |q_{j}| \, . 
\end{align*}

\begin{figure}[h!]
\includegraphics[width=10cm]{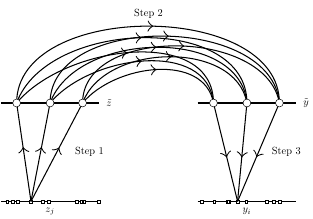}
\caption{An illustration of the algorithm. 
Both the intervals on the left are $(z_{0},z_{0}+R)$, and both the intervals 
on the right are $(y_{0},y_{0}+R)$. 
In the lower intervals, the white squares denote the locations $z_{j}$ and $y_{i}$, and in the upper
intervals the white circles indicate the locations of the equispaced nodes $\tz_{i}$, and $\tz_{j}$.
The arrow illustrate how a point $z_{j}$ communicates with a point $y_{i}$.}
\label{fig:fftcodes}
\end{figure}

A direct computation of $\phi(y_1),\dots, \phi(y_M)$ requires
$O(M\cdot N)$ operations.
On the other hand, the values $\tp(y_{i})$, $i=1,2,\ldots M$,
can be computed in $O((M+N)\cdot p + p^2)$ operations as follows. 
Using~\cref{eq:kerinterp}, $\tp(y_{i})$ can be rewritten as
\begin{align*}
\tp(y_{i}) &= \sum_{j=1}^{N} \sum_{\ell=1}^{p} \sum_{m=1}^{p}
K(\ty_{\ell},\tz_{m}) 
L_{\ell,\ty}(y_{i}) 
L_{m,\tz}(z_{j})q_{j} \, , \\
&= \sum_{\ell=1}^{p} 
L_{\ell,\ty}(y_{i})  
\left(
\sum_{m=1}^{p} 
K(\ty_{\ell},\tz_{m}) 
\left( 
\sum_{j=1}^{N} 
L_{m,\tz}(z_{j})q_{j}  
\right)
\right) \, ,
\end{align*}
for $i=1,2,\ldots M$.
The values $\tp(y_{1}),\dots,\tp(y_M)$, are computed in three steps.
\begin{itemize}
\item {\bf Step 1}: Compute the coefficients $w_{m}$ defined by the formula
\beq
\nonumber
w_{m} = \sum_{j=1}^{N} 
L_{m,\tz}(z_{j})q_{j} \, ,  
\eeq
for each $m=1,2,\ldots p$. 
This step requires $O(N\cdot p)$ operations.
\item {\bf Step 2}: Compute the values $v_{\ell}$ 
at the interpolation nodes
$\ty_{\ell}$ defined by the formula 
\beq
\nonumber
v_{\ell} = \sum_{m=1}^{p}
K(\ty_{\ell},\tz_{m}) w_{m} 
\eeq
for all $\ell=1,2,\ldots p$. This step requires $O(p^2)$ operations.
\item {\bf Step 3}: Evaluate the potential $\tp(y_{i})$ using 
the formula
\beq
\nonumber
\tp(y_{i}) = 
\sum_{\ell=1}^{p} 
L_{\ell,\ty}(y_{i}) v_{\ell} \, ,
\eeq
for all $i=1,2\ldots M$. This step requires $O(M\cdot p)$ operations.
\end{itemize}
See~\cref{fig:fftcodes} for an illustrative figure of the above procedure. 

\subsection{Algorithm}
In this section, we present the main algorithm for the rapid evaluation of the repulsion forces~\cref{eq:nbody2}.
The central strategy is to use piecewise polynomial interpolants of the kernel with equispaced points, 
and use the procedure described in Section \S \ref{subsec:repelprelim}. 

Specifically, suppose that the points $y_{i}$, $i=1,2,\ldots N$ are all contained in the interval
$[\ymin,\ymax]$.
We subdivide the interval $[\ymin,\ymax] = \bigcup_{i=1}^{\Nint} I_{j}$, 
into $\Nint$ intervals of equal length.
Let $\ty_{j,\ell}$
denote $p$ equispaced nodes on the interval $I_{\ell}$ given by
\beq
\label{eq:equinodes}
\ty_{j,\ell} = h/2 + ((j-1) + (\ell-1)\cdot p) \cdot h  \, ,
\eeq
where $h=1/(\Nint \cdot p)$, 
$j=1,2\ldots p$, and $\ell=1,2\ldots \Nint$.
\begin{remark}
The nodes $\ty_{j,\ell}$, $j=1,2\ldots p$, and $\ell=1,2,\ldots \Nint$, defined in~\cref{eq:equinodes}, are also equispaced on the whole interval
$[\ymin,\ymax]$.
\end{remark}

The interaction between any two intervals $I$, $J$, i.e.
$$
\sum_{y_{j} \in J} K(y_{i},y_{j})q_{j} \, , \quad y_{i} \in I
$$
can be accelerated via the algorithm discussed in~\cref{subsec:repelprelim}.
This procedure amounts to using a piecewise polynomial interpolant
of the kernel $K(y,z)$ on the domain $y,z \in [\ymin, \ymax]$ as opposed
to using an interpolant on the whole interval.
We summarize the procedure below.

\begin{itemize}
\item {\bf Step 1}: For each interval $I_{\ell}$, $\ell=1,2,\ldots \Nint$, 
compute the coefficients $w_{m,\ell}$ defined by the formula
\beq
\nonumber
w_{m,\ell} = \sum_{y_{j} \in I_{\ell}} 
L_{m,\ty^{\ell}}(y_{j})q_{j} \, ,  
\eeq
for each $m=1,2,\ldots p$. 
This step requires $O(N\cdot p)$ operations.

\item {\bf Step 2}: 
Compute the values $v_{m,n}$ 
at the equispaced nodes $\ty_{m,n}$ defined by the formula 
\beq
\label{eq:equikereval}
v_{m,n} = \sum_{j=1}^{\Nint} \sum_{\ell=1}^{p}
K(\ty_{m,n},\ty_{\ell,j}) w_{\ell,j} 
\eeq
for all $m=1,2,\ldots p$, $n=1,2\ldots \Nint$. 
This step requires $O((\Nint \cdot p)^2)$ operations.
\item {\bf Step 3}: 
For each interval $I_{\ell}$, $\ell=1,2,\ldots \Nint$, 
compute the potential $\phi(y_{i})$  via the formula
\beq
\nonumber
\phi(y_{i}) = 
\sum_{j=1}^{p} 
L_{j,\ty^{\ell}}(y_{i}) v_{j,\ell} \, ,
\eeq
for all points $y_{i} \in I_{\ell}$.
This step requires $O(N\cdot p)$ operations.
\end{itemize}
In this procedure, the functions $L_{j,\ty^{\ell}}$, $j=1,2,\ldots p$, are the Lagrange polynomials corresponding
to the equispaced interpolation nodes on interval $I_{\ell}$.

In Step 2 of the above procedure, we are evaluating $N-$body interactions on equispaced grid points.
For notational convenience, we rewrite the sum~\cref{eq:equikereval}
\beq
\label{eq:equikereval2}
v_{i} = \sum_{j=1}^{\Nint \cdot p} K(\ty_{i},\ty_{j}) w_{j} \, ,
\eeq
$i=1,2,\ldots \Nint \cdot p$. 
The kernels of interest ($K_{1}$ and $K_{2}$ defined in~\cref{eq:kers})
are translationally-invariant, i.e., the kernels satisfy 
$K(y,z) = K(y+\delta, z+\delta)$ for any $\delta$.
The combination of using equispaced points, along with the
translational-invariance of the kernel, implies that the matrix
associated with the evaluation of the sums~\cref{eq:equikereval2} is Toeplitz.
This computation can thus be accelerated via the 
fast-Fourier transform (FFT), which
reduces the computational complexity of evaluating the 
sums~\cref{eq:equikereval2} from
$O((\Nint \cdot p)^2)$ operations to $O(\Nint \cdot p \log{(\Nint \cdot p)})$.

Algorithm~\ref{alg2} describes the fast algorithm for evaluating the repulsive forces~\cref{eq:nbody2} in  one dimension (s=1)
which has computational complexity
$O( N\cdot p + (\Nint \cdot p) \log{(\Nint \cdot p)})$ .

\begin{algorithm}[!ht]
	\caption{FFT-accelerated Interpolation-based t-SNE (FIt-SNE)\label{alg2}}
  \KwIn{ Collection of points $\{ y_{i} \}_{i=1}^{N}$, source strengths 
  $\{ q_{i}\}_{i=1}^{N}$,
  number of intervals $\Nint$, number of interpolation points per 
  interval $p$}
  
\KwOut{$\phi(y_{i}) =\sum_{j=1}^{N} K(y_{i},y_{j})q_{j}$ for $i=1,2\ldots N$}

For each interval $I_{\ell}$, form the equispaced nodes 
$\ty_{j,\ell}$, $j=1,2,\ldots p$ given by~\cref{eq:equinodes}

\For{$I\leftarrow 1$ \KwTo $\Nint$}
{Compute the coefficients $w_{m,\ell}$ given by 
$$
w_{m,\ell} = \sum_{y_{i} \in I_{\ell}} 
L_{m,\ty^{\ell}}(y_{i})q_{i} \, ,  
$$
$m =1,2,\ldots p$.
}

Use the fast-Fourier transform to compute the values of $v_{m,n}$ given by
\beq
\bbmat
v_{1,1} \\
v_{2,1} \\
\vdots \\
v_{p-1,\Nint}\\
v_{p,\Nint}
\ebmat
= 
\tK \cdot
\bbmat
w_{1,1} \\
w_{2,1} \\
\vdots \\
w_{p-1,\Nint}\\
w_{p,\Nint}
\ebmat \, , 
\eeq
where $\tK$ is the Toeplitz matrix given by
\beq
\tK_{i,j} = K(\ty_{i},\ty_{j}) \, ,
\eeq
$i,j= 1,2,\ldots \Nint \cdot p$. 

\For{$I\leftarrow 1$ \KwTo $\Nint$}
{
Compute $\phi(y_{i})$ at all points $y_{i} \in I_{\ell}$ via
$$
\phi(y_{i}) = \sum_{j=1}^{p} 
L_{j,\ty^{\ell}}(y_{i}) v_{j,\ell}
$$
}
\end{algorithm}

\subsection{Optimal choice of $p$ and $\Nint$}
Recall that the computational complexity of Algorithm~\ref{alg2} is
$O(N \cdot p + \Nint \cdot p \log{(\Nint \cdot p)})$.
We remark that the choice of the parameters $\Nint$ and $p$ depends solely on
the specified tolerance $\eps$ and is independent of the number of points $N$.
Generally, increasing $p$ will reduce the number of 
intervals $\Nint$ required to obtain the same accuracy in the computation.
However, we observe that the reduction in $\Nint$ for an increased $p$ 
is not advantageous from a computational perspective---since, as
the number of points $N$ increases, the computational cost is independent of $\Nint$ and is only 
a function of $p$. 
Moreover, for the t-SNE kernels $K_{1}$ and $K_{2}$ defined in~\cref{eq:kers},
it turns out that for a fixed accuracy the product $\Nint \cdot p$ remains
nearly constant for $p\geq 3$.
Thus, it is optimal to use $p=3$ for all t-SNE calculations.
In a more general environment, when higher accuracy is required and for other translationally invariant
kernels $K$, 
the choice of the number of nodes per interval $p$ and the total number of intervals $\Nint$ can be optimized based on
the accuracy of computation required.
\begin{remark}
Special care must be taken when increasing $p$ in order to achieve higher accuracy due to the Runge phenomenon
associated with equispaced nodes. 
In fact, the kernels that arise in t-SNE are archetypical examples of this
phenomenon.
Since we use only low-order piecewise polynomial interpolation ($p=3$),
we encounter no such difficulties. 
\end{remark}

\subsection{Extension to two dimensions}
The above algorithm naturally extends to two-dimensional embeddings (s=2).
In this case, we divide the computational square $[\ymin, \ymax] \times
[\ymin,\ymax]$ into a collection of $\Nint \times \Nint$ squares
with equal side length, and 
for polynomial interpolation, we use tensor product $p \times p$ 
equispaced nodes on each square. 
The matrix $\tK$ mapping the coefficients $w$ to the coefficients $v$ which is of size
$(\Nint \cdot p)^2 \times (\Nint \cdot p)^2$, is not a Toeplitz matrix, however, it 
can be embedded into a Toeplitz matrix of twice its size. 
The computational complexity of the algorithm analogous to Algorithm~\ref{alg2}
for two-dimensional t-SNE
is $O(N\cdot p^{2} + (\Nint \cdot p)^{2} \log{(\Nint \cdot p)})$.

\subsection{Experiments}
\begin{figure}[!h]
	\includegraphics[width=0.8\textwidth]{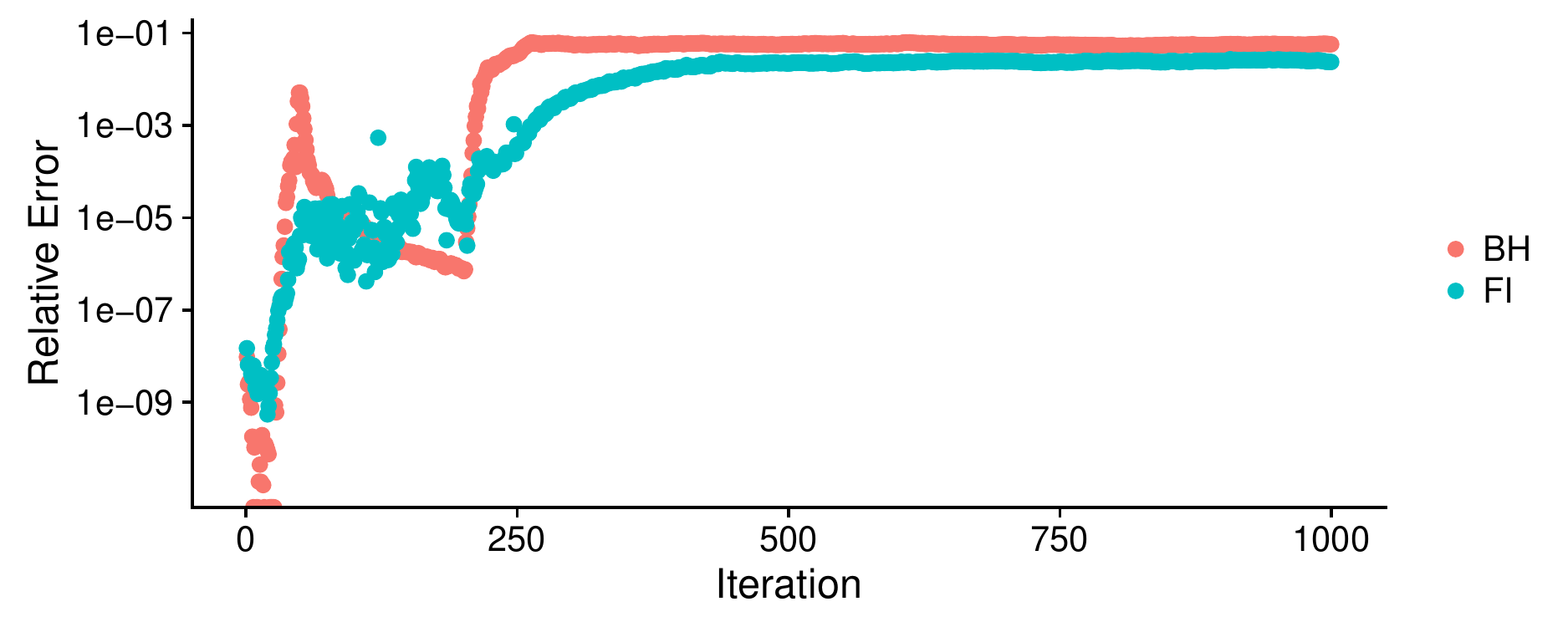}
	\caption{ Accuracy of computing $F_{\text{rep},i}$ using FFT-accelerated Interpolation-based (FI) t-SNE as compared to the
	Barnes-Hut (BH) t-SNE implementation over 1000 iterations (points with error less than $10^{-12}$ are not shown).  }
	\label{fig:accuracy}
\end{figure}

In order to compare the computation time for computing $F_{\text{rep},i}$ using
FFT-accelerated Interpolation-based t-SNE (FIt-SNE) and the Barnes Hut (BH)
implementation t-SNE, we set $p=3$, and chose $\Nint$ to be at either $20$ or
$(\max{y_i} - \min{y_i})$, whichever is larger, so that the accuracy in 2D is
comparable to that of the Barnes-Hut method (with $\theta=0.5$, the default)
for all iterations (Fig. \ref{fig:accuracy}). After the early exaggeration
phase ($\alpha=12$ for the first $200$ iterations), the points expand abruptly,
resulting in decreased accuracy. 

\begin{figure}[h]
	\includegraphics[width=0.8\textwidth]{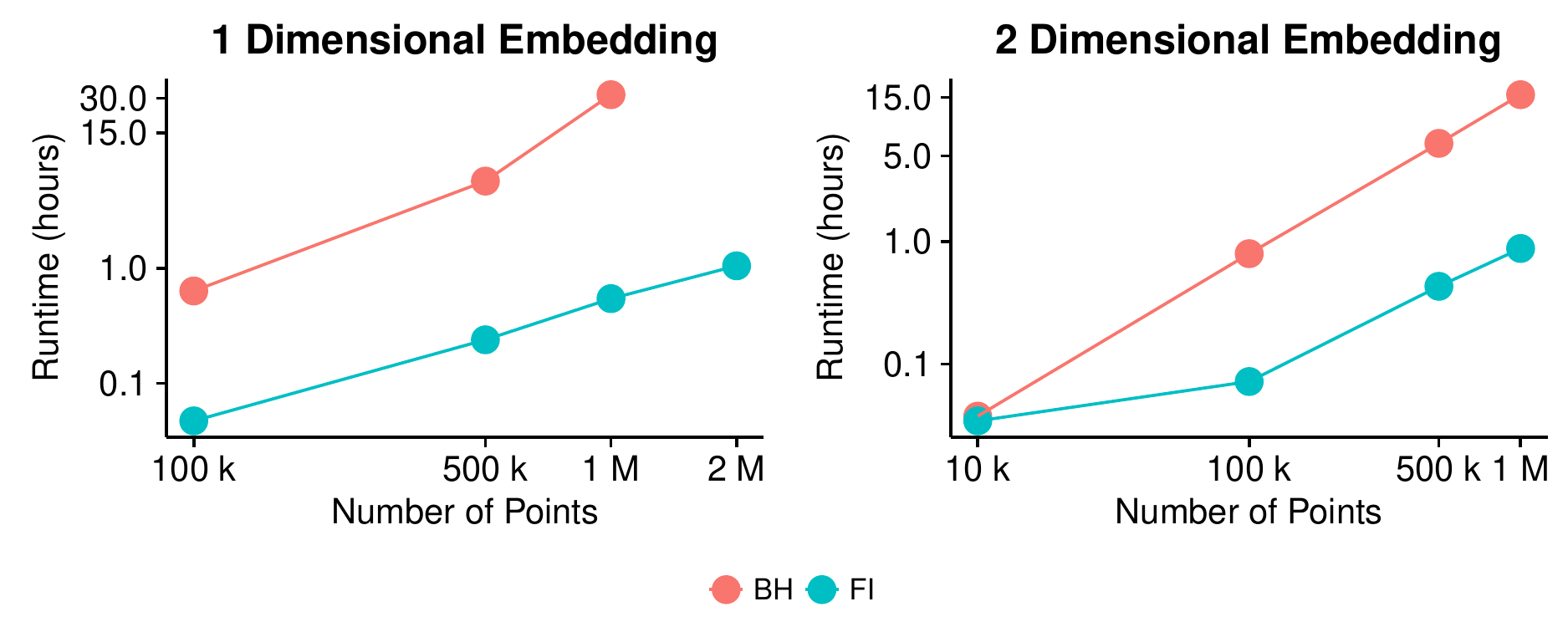}
	\caption{Time required to compute gradients for 1000 iterations of t-SNE using FFT-accelerated Interpolation-based (FI) t-SNE  as compared to the Barnes-Hut (BH) t-SNE implementation. }
	\label{fig:nbody_bench}
\end{figure}

The computation time for computing the gradient for $1000$ iterations, with an
increasing number of points, is shown in Fig. \ref{fig:nbody_bench}. For 1
million points, our method is 15 and 30 times faster than BH when embedding in
1D and 2D respectively, allowing for t-SNE of large datasets on the order of
millions of points.

\section{The attractive forces $F_{\text{attr},i}$}
\label{sec:attr} At each step of gradient descent, the attractive forces on the
$i$th point $$F_{\text{attr},i}=\sum_{j\neq i} {p_{ij} q_{ij} Z (  y_i -  y_j)}$$
attract it to other points in the embedding that are close in the original
space.  In practice, computing the interaction energies $p_{ij}$ between all
pairs of points is too expensive and hence \cite{van2014accelerating} restricts to
computing, for every point, only the interactions with the $k$ nearest
neighbors. In that implementation, nearest neighbors are computed with
vantage-point trees (\cite{yianilos1993data}), which are highly effective in
low dimensions, but are prohibitively expensive when embedding large, high
dimensional datasets.

\begin{figure}[h!]
	\includegraphics[width=0.8\textwidth]{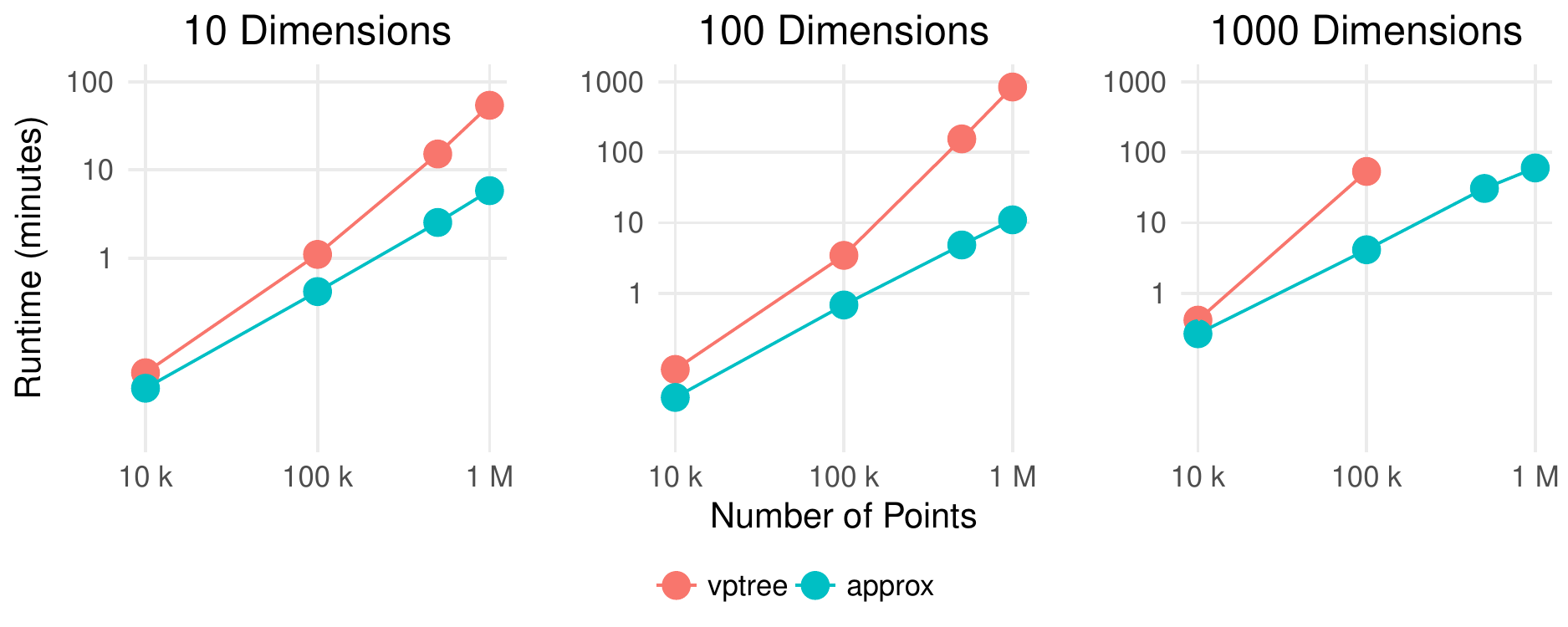}
	\caption{Computation of $p_{ij}$ in first phase of t-SNE}
	\label{fig:IS_bench}
\end{figure}

A recent theoretical advance by \cite{linderman2017randomized} can be used to
optimize this step: it suggests that connecting every point to its
(for example) $k=100$ nearest neighbors is not more effective than connecting
every point to 2 randomly chosen points out of its 100 nearest neighbors.  The
main reason is that this randomized procedure, when executed on point clouds
lying on manifolds, creates expander graphs at the local scale which represent
the local geometry accurately at a slightly coarser level. In the purely
discrete case, this relates to problems in random graphs first raised by Ulam
and Erd\H{o}s-Renyi, and we refer to \cite{linderman2017randomized} for
details.  This simple insight may allow for a massive speedup since the number
of interaction terms $\# \left\{ p_{ij}: p_{ij} \neq 0 \right\}$ is much
smaller.  In practice, we use this result to justify the replacement of nearest
neighbors with approximate nearest neighbors. Specifically, we compute
approximate nearest neighbors using a randomized nearest neighbor method called
ANNOY (\cite{ANNOY}), as we expect the resulting ``near neighbors'' to capture the local
geometry at least as effectively as the same number of nearest neighbors.  We
further accelerated this step by parallelizing the neighbor lookups. The
resulting speed-ups over the vantage point tree approach for computing
$F_{\text{attr},i}$ are shown in Fig. \ref{fig:IS_bench}, as measured on a
machine with 12 Intel Xeon E7540  CPUs clocked at 2.00GHz.

\section{Early and Late Exaggeration}
\label{sec:exaggeration}
In the expression for the gradient descent, the sum of attractive and repulsive
forces, 

$$ \frac{1}{4} \frac{\partial C}{\partial  y_i } =  \alpha \sum_{j\neq i}
{p_{ij} q_{ij} Z (  y_i -  y_j)} - \sum_{j\neq i}{ q_{ij}^2 Z ( y_i - y_j)},$$ 

the numerical quantity $\alpha > 0$ plays a substantial role as it determines
the strength of attraction between points that are similar (in the sense of
pairs $x_i, x_j$ with $p_{ij}$ large). In early exaggeration, first $\alpha =
12$ for the first several hundred iterations, after which it set to $1$ (see
\cite{maaten2008visualizing}). One of the main results of
\cite{linderman2017clustering} is that $\alpha$ plays a crucial role and that
when it is set large enough, t-SNE is guaranteed to separate well-clustered
data and also successfully embed various synthetic datasets (e.g. a swiss
roll) that were previously thought to be poorly embedded by t-SNE.
\begin{figure}[h!]
	\includegraphics[width=\textwidth]{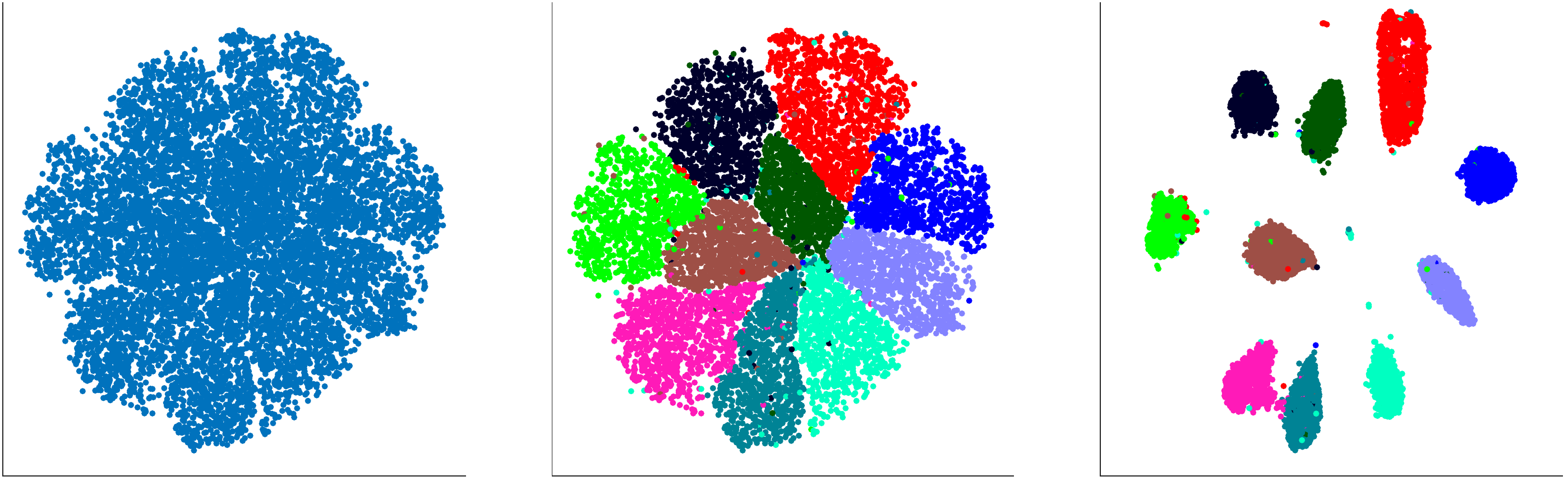}
	\caption{t-SNE embedding of 1 million digits from the Infinite MNIST
	dataset. Late exaggeration of $\alpha =12$ for the last 250 of the 1000
	iterations (right) allows for clusters to be more easily distinguished
	than without it (left and middle).  }
        \label{fig:late_exag}
\end{figure}
We present a novel variation called ``late exaggeration,'' which refers to
setting $\alpha >1$ for the last several hundred iterations. This approach
seems to produce more easily interpretable visualizations: one recurring issue
with t-SNE outputs (see Fig. \ref{fig:late_exag}) is that the arising structure,
while clustered, has its clusters close to each other and does not allow for an
easy identification of segments. By increasing $\alpha$ (and thus the
attraction between points in the same cluster), the clusters contract and are more easily distinguishable. 

\section{t-SNE Heatmaps}
\label{sec:heatmaps}
\begin{figure}[h!]
    \centering
    \begin{minipage}{0.65\textwidth}
        \centering
        \includegraphics[width=1\textwidth]{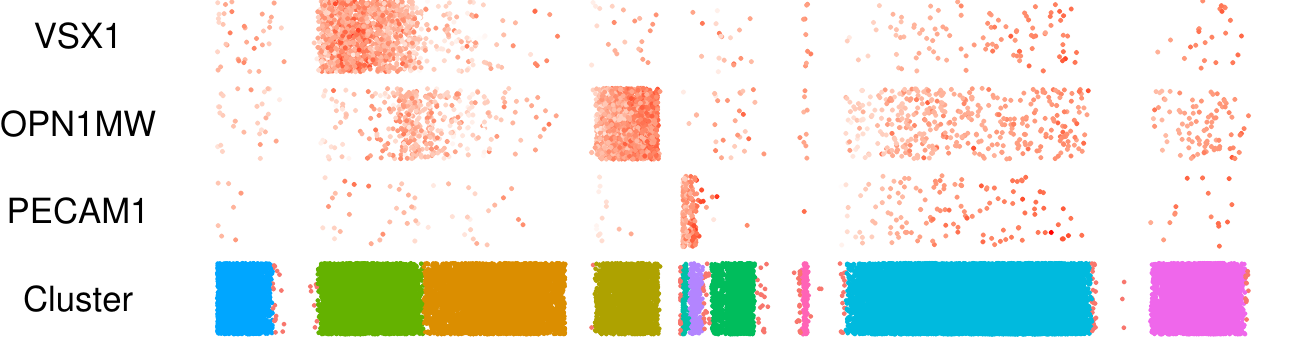} 
    \end{minipage}\hfill
    \begin{minipage}{0.35\textwidth}
        \centering
        \includegraphics[width=1\textwidth]{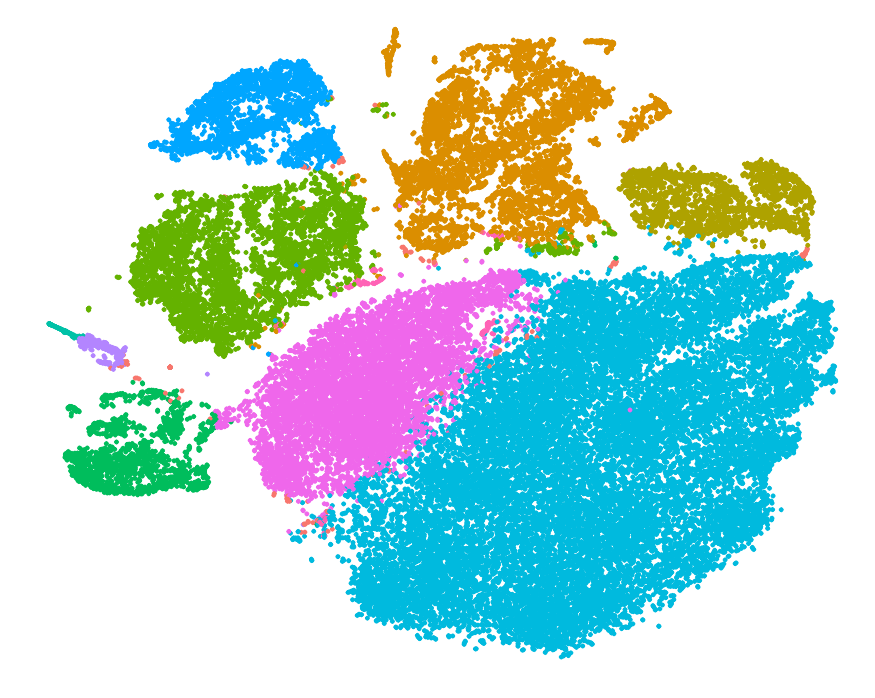} 
    \end{minipage}
	\caption{Left: 1D t-SNE of 49k retinal cells. The first three rows are
	colored by expression of three genes specific to individual clusters,
	and the fourth row is colored by clusters identified using dbscan.
	Uniform noise was added as a “Y-axis” of each row so that more cells
	would be visible. Right: 2D t-SNE plot colored by the clusters assigned
	using the 1D t-SNE, showing that 1D t-SNE contains generally the same
	information as 2D.}
	\label{fig:1d2dtsne}
\end{figure}

The 2D t-SNE plot has become a staple of many scRNA-seq analyses, in
which it is used to visualize clusters of cells, colored by the expression of
interesting genes. Although this information is presented in 2D, users are most
interested in which genes are associated with which clusters, not the 2D shape
or relations of the clusters. In general, the location of clusters with respect
to one another is meaningless, and their 2D shape is not interpretable
and dependent on initialization (\cite{wattenberg2016use}). We hypothesize that 1D t-SNE would contain the
same information as 2D t-SNE, and since it is much more compact, it would allow
simultaneous visualization of the expression of hundreds of genes in a
heatmap-like fashion. The general idea is shown in Fig. \ref{fig:1d2dtsne}, where we embedded
the 49k retinal cells of \cite{macosko2015highly} using 1D and 2D t-SNE and assigned corresponding
points the same color to show that the embeddings are equivalent. 

\begin{figure}[h!]
        \includegraphics[width=\textwidth]{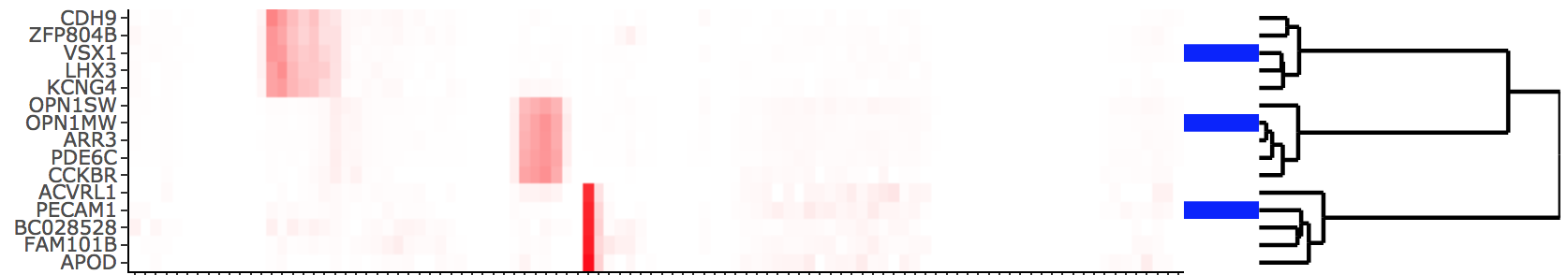}
	\caption{t-SNE heatmap using 1D embedding of Fig. \ref{fig:1d2dtsne}.}
        \label{fig:tsne_heatmap}
\end{figure}

This 1D t-SNE representation can be extended into a visualization we call
``t-SNE Heatmaps.'' The 1D t-SNE is first discretized into $p$ bins, and we sum
the expression of each gene in each of the bins, such that each gene $g$ is a
vector in $\mathbb{R}^p$. A distance between genes is now defined as the
Euclidean distance between these vectors. In practice, the user provides a set
of genes of interest (GOI), and this gene set is then enriched with genes that
are closest to the genes of interest in this metric. Each of the $p$-vectors
corresponding to this new gene set are rows in a heatmap, and can be used to
visualize hundreds of genes' expression on the t-SNE embedding. We give a small
example in Fig. \ref{fig:tsne_heatmap}, where three genes in corresponding to
known  Retinal subpopulations are enriched with the four genes closest to each
in the t-SNE metric and visualized in this heatmap format. 

In general, the rows
need not correspond to individual genes; if a method for clustering genes is
available, then t-SNE Heatmaps can be used to visualize how the cell
clusters are associated with gene clusters. Overall, t-SNE Heatmaps uses
the 1D t-SNE of the cells to define a distance on the genes, and
could be similarly extended to define a distance on the cells. This would allow
for t-SNE based iterative methods to be developed, similar to \cite{mishne2017data},
where the the embedding of the cells is used to improve the embedding of the
genes, which is then used to improve the embedding of the cells, and so forth.
\section{Out-of-Core PCA}
\label{sec:oocPCA}
The methods for t-SNE presented above allow for embedding of millions of points
in only hours, but can only be used to reduce the dimensionality of datasets
that can fit in the memory. For many large, high dimensional datasets,
specialized servers must be used in order to simply load the data. For
instance, a single cell RNA-seq dataset with a million cells, where the
expression of $20,000$ genes are measured for each cell, requires 160GB of
memory - far exceeding the capacity of a standard personal computer. In order
to allow for visualization and analysis of such datasets on resource-limited
machines, we present an out-of-core implementation of randomized PCA, which can
be used to compute the top few  (e.g. 50) principal components of a dataset to
high accuracy, without ever loading it in its entirety (\cite{halko2011algorithm}).
\subsection{Randomized Methods for PCA}
The goal of PCA is to approximate the matrix being analyzed (after mean
centering of its columns) with a low-rank matrix. PCA is primarily useful when
such an approximation makes sense; that is, when the matrix being analyzed is
approximately low-rank. If the input matrix is low-rank, then by definition,
its range is low-dimensional. As such, when the input matrix is applied to a
small number of random vectors, the resulting vectors nearly span its range.
This observation is the core idea behind randomized algorithms for PCA:
applying the input matrix to a small number of random vectors results in
vectors that approximate the range of the matrix. Then, simple linear algebra
techniques can be used to compute the principal components. Notably, the only
operations involving the large input matrix are matrix-vector multiplications,
which are easily parallelized, and for which highly optimized implementations
exist. Randomized algorithms have been rigorously proven to be remarkably
accurate with extremely high probability (e.g.
\cite{halko2011finding,witten2015randomized}), because for a rank-$k$ matrix,
as few as $l=k+2$ random vectors are sufficient for the probability of missing a
significant part of the range to be negligible. The algorithm and its
underlying theory are covered in detail in \cite{halko2011finding}. An
easy-to-use ``black box'' implementation of randomized PCA is available and
described in  \cite{li2017algorithm}, but it requires the entire matrix to be
loaded in the memory. We present an out-of-core implementation of PCA in R,
oocPCA, allowing for decomposition of matrices which cannot fit in the
memory.

\begin{algorithm}[ht]
	\caption{Out-of-Core PCA (oocPCA)\label{alg}}
  \KwIn{Matrix $A$ of size $m \times n$ stored in slow memory, non-negative integers $its$, $k$, $l$, $b$, where $0 <k \leq l < \min(m,n)$, and $l$ defaults to $k+2$ }
\KwOut{Orthonormal $U$ of size $m \times k$, non-negative diagonal matrix $\Sigma$ of size $k \times k$, orthonormal $V$ of size $n \times
k$, such that  $A \approx U \Sigma V^* $}

Generate uniform random matrix $\Omega$ of size $n \times l$

Form $Y_0 = A \Omega$ block-wise, $b$ rows at a time

Renormalize with LU factorization $L_0U_0 = Y_0$

\For{$i\leftarrow 1$ \KwTo $its$}{
Form $Y_i = AA^* L_{i-1}$ block-wise, $b$ rows at a time

\If{$i < its$}{
Renormalize with LU factorization $L_{i}U_{i} = Y_i$}
}

Renormalize with QR factorization $QR = Y_i$

Compute SVD of small matrix $U' \Sigma V^* = Q^* A$

Set $U = QU'$

\end{algorithm}

\subsection{Implementation}
Our implementation is described in Algorithm 1. Given an $m \times n$ matrix of
doubles $A$, stored in row-major format on the disk of a machine with $M$ bytes
of available memory, the number of rows that can fit in the memory is calculated as $b =
\left\lfloor\frac{M}{8mn} \right\rfloor$. The only operations performed using
$A$ are matrix multiplications, which can be performed block-wise.
Specifically, the matrix product $AB$, where $B$ is an $n\times p$
matrix stored in the fast memory, can be computed by loading the first $b$ rows
of $A$, and forming the inner product of each row with the columns of $B$. The
process can be continued with the remaining blocks of the matrix, essentially
``filling in'' the product $AB$ with each new block. In this manner, left
multiplication by $A$ can be computed without ever loading the full matrix $A$.

By simply replacing the matrix multiplications in \cite{li2017algorithm}'s
implementation with block-wise matrix multiplication, an out-of-core algorithm
can be obtained. However, significant optimization is possible. The run-time of
an out-of-core algorithm is almost entirely determined by disk access time;
namely, the number of times the matrix must be loaded to the memory. As
suggested in \cite{li2017algorithm}, the renormalization step between the
application of $A$ and $A^*$ is not necessary in most cases, and in the
out-of-core setting, doubles the number of times $A$ must be loaded per power
iterations. In our implementation, we remove this renormalization step, and
apply $AA^*$ simultaneously, hence requiring the matrix only be loaded once per
iteration.

Our implementation is in C++ with an R (\cite{R}) wrapper. For maximum
optimization of linear algebra operations, we use the highly parallelized Intel
MKL for all BLAS functions (e.g. matrix multiplications). The R wrapper
provides functions for PCA of matrices in CSV and in binary format.
Furthermore, basic preprocessing steps including $\log$ transformation and mean
centering of rows and/or columns can also be performed prior to decomposition, so that
the matrix need not ever be fully stored in the memory.
\subsection{Experiments}
\begin{figure}[h!]  \label{fig:decay}
\begin{center}
\begin{tabular}{ |c|c|c|c|c|c|c| }
	\hline
	Memory (GB) & 1    & 2    & 8    & 32    & 128 & 300\\
	\hline
	Time (Min) & 15.9 & 12.8 & 12.7 & 12.0 & 10.5 & 8.4\\
	\hline
\end{tabular}
\end{center}
\caption{PCA of $\num[group-separator={,}]{1000000} \times \num[group-separator={,}]{30000}$ rank-$50$ matrix with varying memory limitations}
\label{fig:ooc_1mill}
\end{figure}

We generated a random $\num[group-separator={,}]{1000000} \times
\num[group-separator={,}]{30000}$ rank-$50$ matrix of doubles, which would
require $240$GB to simply store in the memory, far exceeding the memory
capacity of a personal computer. Using oocPCA we can compute the top
principal components of the matrix with much less memory (Fig. \ref{fig:ooc_1mill}).  By storing only 1GB
of the matrix in the memory at a time, and all other parameters set to default,
the top $50$ principal components of this matrix can be computed in 16
minutes, while attaining an approximation accuracy of $\sim 10^{-9}$ in the
spectral norm. 
\section{Summary and Discussion}
In this work, we present an implementation of t-SNE that allows for embedding
of high dimensional datasets with millions of points in only a few hours. Our
implementation includes a ``late exaggeration'' feature, which can make it easier to
identify clusters in t-SNE plots. Finally, we presented an out-of-core
algorithm for PCA, allowing for analysis and visualization of datasets that
cannot fit into the memory. A natural extension of the present work is the
analysis of large scRNA-seq datasets, and we are currently applying this
approach to analyze the 1.3 million mouse brain cells dataset of \cite{10X}. Our
methods allow for visualization of these datasets, without subsampling, on a
standard personal computer in a reasonable amount of time.
\section{Software Availability}
FIt-SNE and ooPCA are both available at \url{https://github.com/KlugerLab/}.  The FIt-SNE repository also contains a script for producing t-SNE Heatmap visualizations. 

\section{Acknowledgements}
The authors would like to thank Vladimir Rokhlin and Mark Tygert for many useful discussions.
\bibliography{tsne_draft_v5}{}

\begin{thebibliography}{}

\bibitem[\protect\astroncite{{10X Genomics}}{2016}]{10X}
{10X Genomics} (2016).
\newblock Transciptional profiling of 1.3 million brain cells with the chromium
  single cell 3' solution.
\newblock {\em Application Note}.

\bibitem[\protect\astroncite{Barnes and Hut}{1986}]{barnes1986hierarchical}
Barnes, J. and Hut, P. (1986).
\newblock A hierarchical {O}({N} log {N}) force-calculation algorithm.
\newblock {\em Nature}, 324(6096):446--449.

\bibitem[\protect\astroncite{Bernhardsson}{2017}]{ANNOY}
Bernhardsson, E. (2017).
\newblock Annoy: Approximate nearest neighbors in c++/python optimized for
  memory usage and loading/saving to disk.
\newblock https://github.com/spotify/annoy.

\bibitem[\protect\astroncite{Halko et~al.}{2011a}]{halko2011algorithm}
Halko, N., Martinsson, P.-G., Shkolnisky, Y., and Tygert, M. (2011a).
\newblock An algorithm for the principal component analysis of large data sets.
\newblock {\em SIAM Journal on Scientific computing}, 33(5):2580--2594.

\bibitem[\protect\astroncite{Halko et~al.}{2011b}]{halko2011finding}
Halko, N., Martinsson, P.-G., and Tropp, J.~A. (2011b).
\newblock Finding structure with randomness: Probabilistic algorithms for
  constructing approximate matrix decompositions.
\newblock {\em SIAM Review}, 53(2):217--288.

\bibitem[\protect\astroncite{Li et~al.}{2017}]{li2017algorithm}
Li, H., Linderman, G.~C., Szlam, A., Stanton, K.~P., Kluger, Y., and Tygert, M.
  (2017).
\newblock Algorithm 971: an implementation of a randomized algorithm for
  principal component analysis.
\newblock {\em ACM Transactions on Mathematical Software (TOMS)}, 43(3):28.

\bibitem[\protect\astroncite{Linderman et~al.}{2017}]{linderman2017randomized}
Linderman, G.~C., Mishne, G., Kluger, Y., and Steinerberger, S. (2017).
\newblock Randomized near neighbor graphs, giant components, and applications
  in data science.
\newblock {\em arXiv preprint arXiv:1711.04712}.

\bibitem[\protect\astroncite{Linderman and
  Steinerberger}{2017}]{linderman2017clustering}
Linderman, G.~C. and Steinerberger, S. (2017).
\newblock Clustering with t-{SNE}, provably.
\newblock {\em arXiv preprint arXiv:1706.02582}.

\bibitem[\protect\astroncite{Macosko et~al.}{2015}]{macosko2015highly}
Macosko, E.~Z., Basu, A., Satija, R., Nemesh, J., Shekhar, K., Goldman, M.,
  Tirosh, I., Bialas, A.~R., Kamitaki, N., Martersteck, E.~M., et~al. (2015).
\newblock Highly parallel genome-wide expression profiling of individual cells
  using nanoliter droplets.
\newblock {\em Cell}, 161(5):1202--1214.

\bibitem[\protect\astroncite{Mishne et~al.}{2017}]{mishne2017data}
Mishne, G., Talmon, R., Cohen, I., Coifman, R.~R., and Kluger, Y. (2017).
\newblock Data-driven tree transforms and metrics.
\newblock {\em IEEE Transactions on Signal and Information Processing over
  Networks}.

\bibitem[\protect\astroncite{{R Core Team}}{2017}]{R}
{R Core Team} (2017).
\newblock {\em R: A Language and Environment for Statistical Computing}.
\newblock R Foundation for Statistical Computing, Vienna, Austria.

\bibitem[\protect\astroncite{van~der Maaten}{2014}]{van2014accelerating}
van~der Maaten, L. (2014).
\newblock Accelerating t-{SNE} using tree-based algorithms.
\newblock {\em Journal of machine learning research}, 15(1):3221--3245.

\bibitem[\protect\astroncite{van~der Maaten and
  Hinton}{2008}]{maaten2008visualizing}
van~der Maaten, L. and Hinton, G. (2008).
\newblock Visualizing data using t-{SNE}.
\newblock {\em Journal of Machine Learning Research}, 9(Nov):2579--2605.

\bibitem[\protect\astroncite{Wattenberg et~al.}{2016}]{wattenberg2016use}
Wattenberg, M., Vi{\'e}gas, F., and Johnson, I. (2016).
\newblock How to use t-{SNE} effectively.
\newblock {\em Distill}, 1(10):e2.

\bibitem[\protect\astroncite{Witten and Candes}{2015}]{witten2015randomized}
Witten, R. and Candes, E. (2015).
\newblock Randomized algorithms for low-rank matrix factorizations: sharp
  performance bounds.
\newblock {\em Algorithmica}, 72(1):264--281.

\bibitem[\protect\astroncite{Yianilos}{1993}]{yianilos1993data}
Yianilos, P.~N. (1993).
\newblock Data structures and algorithms for nearest neighbor search in general
  metric spaces.
\newblock In {\em SODA}, volume~93, pages 311--321.

\end{thebibliography}
\bibliographystyle{apa}

\section{Appendix}
In Section \S \ref{sec:repul} we noted that the repulsive forces $F_{\text{rep},k}$ defined in~\cref{eq:frep} 
can be expressed as $s+2$ 
sums of the form
\begin{equation*}
\phi(y_{i}) = \sum_{j=1}^{N} K(y_{i},y_{j}) q_{j} 
\end{equation*}
where the kernel $K(y,z)$ is either
\begin{equation*}
K_{1}(y,z) = \frac{1}{(1+\| y-z \|^2)} \, , \quad \text{or} \quad K_{2}(y,z) = \frac{1}{(1+\|y-z \|^2)^2} \, , 
\end {equation*}
	for $y,z \in \bR^{s}$. In this appendix, we demonstrate for $s=2$. The following $4$ sums are computed at each step of gradient descent:

\begin{align*}
	h_{1,j} &= \sum_{\substack{\ell = 1 \\ \ell \neq j}}^{N}
\frac{1}{(1 + \| y_\ell - y_j\|^2)} \, ,  \\
	h_{2,k} & = \sum_{\substack{\ell=1\\ \ell \neq k}}^{N}
	\frac{y_\ell(1)}
{\left(1 + \| y_\ell - y_k\|^2\right)^{2}} \, ,\\
	h_{3,k} &= \sum_{\substack{\ell=1 \\ \ell \neq k}}^{N} 
	\frac{y_\ell(2)}
{\left(1 + \| y_\ell - y_k\|^2\right)^{2}} \, , \\
	h_{4,k} &= \sum_{\substack{\ell=1 \\ \ell \neq k}}^{N} 
\frac{1}
{\left(1 + \| y_\ell - y_k\|^2\right)^{2}} \, .
\end{align*}

At each step of gradient descent, the repulsive forces can then be expressed in terms of these
$4$ sums as follows:
\begin{align*}
	F_{\text{rep},k}(1) &= 
\left( 
\sum_{\substack{\ell=1 \\ \ell \neq k}}^{N} 
\frac{y_\ell(1) - y_k(1)}
{\left(1 + \| y_\ell - y_k\|^2\right)^{2}} \right) \Bigg/ 
\left(
\mathop{
\sum_{\substack{j = 1}}^{N}
\sum_{\substack{\ell = 1}}^{N}}_{\ell \neq j}
\frac{1}{(1 + \| y_\ell - y_j\|^2)}\right) 
\\
	&=\left(h_{2,k} - y_k(1)h_{4,k}\right) /Z,\\
	F_{\text{rep},k}(2) &= 
\sum_{\substack{\ell=0 \\ \ell \neq k}}^{N} 
\frac{y_\ell(2) - y_k(2)}
{\left(1 + \| y_\ell - y_k\|^2\right)^{2}} \Bigg/
\left(
\mathop{
\sum_{\substack{j = 1}}^{N}
\sum_{\substack{\ell = 1}}^{N}}_{\ell \neq j}
\frac{1}{(1 + \| y_\ell - y_j\|^2)}\right) 
\\
	&=\left(h_{3,k} - y_k(2)h_{4,k}\right)/Z,\\
\end{align*}  
where 
	\begin{align*}
	Z &= \sum_{\substack{j = 1}}^{N} h_{1,j} \, .
	\end{align*}

\end{document}